**Note: This preprinted paper was accepted in IEEE Access as follows.**

Modified Manuscript Title: ***Machine Learning Evaluation Metric Discrepancies across Programming Languages and Their Components in Medical Imaging Domains: Need for Standardization***

# Machine Learning Evaluation Metric Discrepancies across Programming Languages and Their Components: Need for Standardization

Mohammad R. Salmanpour[1,2,3*#], Morteza Alizadeh[3,4#], Ghazal Mousavi[3,5], Saba Sadeghi[3,6], Sajad Amiri[3], Mehrdad Oveisi[3,7], Arman Rahmim[1,2], Ilker Hacihaliloglu[1,8]

[1] Department of Radiology, University of British Columbia, Vancouver, BC, Canada
[2] Department of Integrative Oncology, BC Cancer Research Institute, Vancouver, BC, Canada
[3]Technological Virtual Collaboration (TECVICO CORP.), Vancouver, BC, Canada
[4] Department of Mathematics, University of Isfahan, Isfahan, Iran
[5] School of Electrical and Computer Engineering, University of Tehran, Tehran, Iran
[6] Department of Statistics, Shiraz University, Shiraz, Iran
[7] Department of Computer Science, University of British Columbia, Vancouver, BC, Canada
[8]Department of Medicine, University of British Columbia, Vancouver, BC, Canada

**Corresponding Author:**
(*) Mohammad R. Salmanpour, PhD
Department of Radiology, University of British Columbia, Vancouver BC, Canada
Email. m.salmanpour@ubc.ca

(#) These authors had equal contributions

## ABSTRACT

This study evaluates metrics for tasks such as classification, regression, clustering, correlation analysis, statistical tests, segmentation, and image-to-image (I2I) translation. Metrics were compared across Python libraries, R packages, and Matlab functions to assess their consistency and highlight discrepancies. The findings underscore the need for a unified roadmap to standardize metrics, ensuring reliable and reproducible ML evaluations across platforms. This study examined a wide range of evaluation metrics across various tasks and found only some to be consistent across platforms, such as (i) Accuracy, Balanced Accuracy, Cohens Kappa, F-beta Score, MCC, Geometric Mean, AUC, and Log Loss in binary classification; (ii) Accuracy, Cohens Kappa, and F-beta Score in multi-class classification; (iii) MAE, MSE, RMSE, MAPE, Explained Variance, Median AE, MSLE, and Huber in regression; (iv) Davies-Bouldin Index and Calinski-Harabasz Index in clustering; (v) Pearson, Spearman, Kendall's Tau, Mutual Information, Distance Correlation, Percbend, Shepherd, and Partial Correlation in correlation analysis; (vi) Paired t-test, Chi-Square Test, ANOVA, Kruskal-Wallis Test, Shapiro-Wilk Test, Welch's t-test, and Bartlett's test in statistical tests; (vii) Accuracy, Precision, and Recall in 2D segmentation; (viii) Accuracy in 3D segmentation; (ix) MAE, MSE, RMSE, and R-Squared in 2D-I2I translation; and (x) MAE, MSE, and RMSE in 3D-I2I translation. Given observation of discrepancies in a number of metrics (e.g. precision, recall and F1 score in binary classification, WCSS in clustering, multiple statistical tests, and IoU in segmentation, amongst multiple metrics), this study concludes that ML evaluation metrics require standardization and recommends that future research use consistent metrics for different tasks to effectively compare ML techniques and solutions.

**Keywords:** ML Evaluation Metrics; Consistency of Evaluation Metrics in Multi-Framework; 2D/3D Medical Images; Evaluation Metric Roadmap.

## 1. INTRODUCTION

Machine learning (ML) have shown significant value in solving complex problems across different fields, including classification, regression, clustering, correlation, segmentation, and image-to-image (I2I) translation [1]. While this study primarily explores healthcare applications, the findings have broader implications, positively impacting not only the healthcare domain but also the larger ML community and other research areas. Moreover, ML algorithms are increasingly adopted, accurately evaluating their performance is crucial for both research advancements and practical applications [2]. Reliable evaluation metrics allow researchers and developers to compare, improve, and choose the best models, drive progress and ensure effective ML solutions [3].

A given programming language, e.g. Python, Matlab and R, offers its own set of components (i.e. libraries, packages, and functions) that provide different ways to calculate the same evaluation metrics. These implementations are often community-driven, which, while fostering innovation and rapid development, might also introduce inconsistencies and errors [4]. Variations in coding standards, documentation quality, and maintenance practices among community-based projects might result in discrepancies in metric calculations, leading to unexpected mistakes and unreliable values [3]. This lack of standardization can result in different outcomes when the same metric is used in different environments, making it difficult to compare models and reducing trust in the results.

These inconsistencies can have significant consequences, e.g. in critical areas such as healthcare, finance, and autonomous driving systems [6, 7]. Inconsistent metric evaluations can lead to poor model selection, incorrect conclusions, and real-world failures. For instance, in clustering and complex statistical analyses, the choice and implementation of evaluation metrics can greatly affect how data is interpreted and the decisions that follow [8, 9, 10]. Similarly, in specialized areas like semantic segmentation and I2I translation, ongoing debates about the best evaluation measures highlight the problems caused by inconsistent metric implementations [11, 12, 13]. Moreover, many studies report improvements based on metrics calculated using different languages, packages, and functions [14, 15]. However, the variability in metric implementations may result in overestimates of model performance, as comparisons across inconsistent metrics are not reliable. This potential overestimation undermines the validity of these studies and highlights the critical need for standardized metric implementations to ensure accurate and comparable results. These issues not only make research harder to reproduce but also slow down advancements in ML by introducing uncertainty and variability in model assessments.

Driven by the need for reliable and consistent ML evaluations, our study conducts a thorough assessment of various evaluation metrics across different programming languages. We examine a wide range of ML tasks, such as classification, regression, clustering, correlation, statistical analysis, segmentation, and I2I translation. Our research demonstrates that differences in how these metrics are implemented across different components can lead to significant variations in the results produced by identical algorithms in different environments. To address this significant gap, this study aims to develop a comprehensive roadmap that systematically aligns and compares existing evaluation metrics across various components such as libraries in Python, functions in Matlab, and packages in R programming languages, tailored to specific tasks. To date, no study has undertaken the task of investigating the consistency of ML evaluation metrics across different programming languages. The present study focused on three widely-used languages in ML—Python, Matlab, and R—due to their extensive libraries, functions, and packages for ML tasks, which are commonly used by ML engineers in practice. This work aims to identify which metrics produce consistent results across various implementations and highlight those that require re-evaluation. By providing this structured overview, we seek to guide researchers and developers in selecting more consistent metrics and promoting the establishment of unified standards. Additionally, our roadmap will serve as a valuable resource for identifying the best practices and areas needing improvement, thereby fostering the development of more robust and trustworthy ML models.

## 2. MATERIAL AND METHODS

### 2.1. Data

We employed multiple medical imaging datasets for different tasks, as shown in Supplemental Table S1, which includes details on dataset size, task type, specific imaging modalities used, etc. D1 and D2 include 1023 deep [16] and 215 handcrafted radiomics features [17], respectively, extracted from PET and CT images of 408 head and neck cancer patients using an autoencoder and standardized PySERA within VISERA software (*visera.ca*) [18]. D3 includes 215 handcrafted radiomics features, extracted from PET images of 408 head and neck cancer patients using the standardized PySERA within VISERA software. D4 and D5 include 1023 deep radiomics features, respectively, extracted from CT and PET images of 199 lung cancer patients using an autoencoder within VISERA software [19] [20]. D6 and D7 include 215 handcrafted radiomics features, extracted from CT and PET images of 199 lung cancer patients using the standardized PySERA within VISERA software. D8 includes 215 handcrafted radiomics features, extracted from MRI images of 740 prostate cancer patients using the standardized PySERA within VISERA software [21].

D9 comprises two features: surface area from the Morphology feature category and dependence counts non-uniformity, selected from the 3D Neighbouring Grey Level Dependence Matrix (NGLDM) within the D8 dataset. Similarly, D10 comprises two features: high dependence high grey level emphasis and grey level non-uniformity, both derived from the 3D-NGLDM within the D8 dataset. D11 comprises two features: Surface to volume ratio from the morphology feature category and low dependence emphasis, selected from the 3D-NGLDM within the D8 dataset. D12 and D13 include a total of 1,027 segmentation masks obtained from CT images and 2D MEDSAM and 3D ReconNet segmentation networks. Specifically, D12 contains 2D masks, while D13 comprises 3D masks. Moreover, D14 and D15 include a total of 740 translated MRI images obtained from ultrasound images applied to 2D CycleGAN and 3D CycleGAN networks, respectively [21]. Specifically, D14 contains 2D MRI slices, while D15 comprises 3D translated MRI images. Finally, we generated 11 different random datasets (with random state 42) for each task for further validation.

### 2.2. ML Algorithms

We employed different algorithms (elaborated in Supplemental Section 1.2) for different tasks, as follows. In binary & multi-class classification, Regression, clustering, 2D-segmentation, 3D-segmentation, 2D-I2I translation, and 3D-I2I we utilized Random Forest Classifier [22], Random Forest Regressor [23], K-Means, 2D MEDSAM, 3D ReconNet [24], 2D CycleGAN, 3D CycleGAN algorithms [21], respectively. Performance of these algorithms was not the focus of this study, as the primary aim was to evaluate ML evaluation metric discrepancies.

**2.3 Study of different evaluation metrics across components in programming languages**

As shown in Table 1, this study employs 88 components including 29 libraries, 31 packages, and 28 functions in Python, R, and Matlab programming languages to calculate different evaluation metrics, respectively. In Python, we used different libraries such as Scikit-Learn, TensorFlow, MLxtend, PyCM, Evaluate, NLTK, Ignit, Imbalanced Learn, StatsModels, SciPy, Pandas, dcor, Numpy, Pingouin, Vaex, Xarray, Spark, Geopandas, Causal Impact, Dask Array, SymPy, ResearchPy, SimpleITK, MedPy, Hausdorff, TorchMetrics, Keras, Monai, and Scikit-Image. In R Language, we utilized packages such as Caret, MLmetrics, Yardstick, Psych, E1071, IRR, MCCR, Cluster, Factoextra, ClusterCrit, Infotheo, Energy, Entropy, Corrplot, Hmisc, Metrics, ModelMetrics, DescTools, Imager, Forecast, HydroGOF, TDR, Rminer, MetricsWeighted, Mclust, Infer, Rstatix, Car, Stats, pROC, and ROCR. In Matlab, we used different functions, such as namely Jaccard, geomean, std, mae, immse, rmse , mean, var, median, silhouette, Rsquared.Ordinary, psnr, ssim, ttest2, ttest, kstest2, chi2gof, anova1, ranksum, swtest, vartest2, ttest, ttest2, vartestn, signrank, chi2cdf, bwperim, and bwdist, as shown in Table1.

As shown in Table 1 and elaborated in Supplemental Section 1.3, the above-mentioned components support ~60 metrics, including:

(i) Binary and (ii) multi-class classification tasks employ evaluation metrics such as Accuracy (Acc), Precision (Pr), Recall (Re), F1 Score (F1), Cohen’s Kappa (CK), Matthew's Correlation Coefficient (MCC), Binary Cross-Entropy (Log Loss, LL), Balanced Accuracy (BAcc), F-beta Score (F-be), Jaccard Index (JI, Similarity Coefficient), AUC, and Geometric Mean (G-mean, G-m).

(iii) Regression tasks include metrics such as Mean Absolute Error (MAE), Mean Square Error (MSE), Root Mean Square Error (RMSE), Mean Absolute Percentage Error (MAPE), R-Squared ($R^2$), Mean Squared Log Error (MSLE), Explained Variation (EV), Median Absolute Error (Med AE), Tweedie Deviance (TD), and Huber (Hu).

(iv) Clustering tasks consist of metrics such as the Silhouette Score (Sil), Davies-Bouldin Index (DaBo), Calinski-Harabasz Index (CaHa), and Within-Cluster Sum of Squares (WCSS).

(v) Correlation tasks include metrics such as Pearson (Pear), Spearman (Spea), Kendall's Tau (Tau), Mutual Information (MI), Distance Correlation (DCorr), Bicor (Bic), Percbend (Perc), Shepherd (Shep), and Partial Corr (PCorr).

(vi) Statistical tests include independent t-test (itt), Paired t-test (ptt), Kolmogorov-Smirnov Test (kst), ANOVA (ANO), Kruskal-Wallis Test (kwt), Mann-Whitney U Test (mwut), Shapiro-Wilk Test (swt), F-test (ft), Welch's t-test (wet), Bartlett's test (bt), Levene's test (lt), Chi-Square Test (chi2), Z-test (zt), Wilcoxon signed-rank test (wsrt), and Permutation test (pt).

(vii) 2D and (viii) 3D segmentation tasks consist of metrics such as Accuracy, Precision, Recall, F1 Score, Dice Coefficient (Dice), Intersection over Union (IoU), Mean IoU, Boundary F1 Score (BF), Hausdorff Distance (Haus), Adapted Rand Error (AdaR), Adjusted Rand Error (AdjR), and Variation of Information (VoI).

(ix) 2D and (x) 3D I2I translation tasks include various evaluation metrics such as MAE, MSE, RMSE, $R^2$, Peak Signal-to-Noise Ratio (PSNR), and Structural Similarity Index Measure (SSIM).

In the present work, we document discrepancies across the above-mentioned evaluation metrics as implemented in components of programming languages.

**Table 1.** Overview of metric availability across programming components in Python, R, and Matlab. This table offers a comprehensive listing of ML evaluation metrics across components in three major programming languages: Python, R, and Matlab. Each row corresponds to a specific metric, grouped by different ML tasks. The Green, orange and blue sections show the availability of each metric in various Python libraries, R packages, and Matlab functions, respectively.

| | | Python | | | | | | | | | | | | | | | | | | | | | | | | | | | | |
|---|---|---|---|---|---|---|---|---|---|---|---|---|---|---|---|---|---|---|---|---|---|---|---|---|---|---|---|---|---|---|
| Tasks | Metrics | Scikit-Learn | TensorFlow | PyTorch | MLxtend | PyCM | Evaluate | NLTK | Ignit | Imbalanced learn | Statsmodels | Scipy | Pandas | Numpy | dcor | Pingouin | Vaex | Xarray | Spark | Geopandas | Causal Impact | Dask Array | SymPy | ResearchPy | SimpleITK | MedPy | Hausdorff | TorchMetrics | Monai | Scikit-Image |
| Classification | Acc | ✻ | ✻ |  | ✻ | ✻ | ✻ | ✻ | ✻ |  | ✻ |  |  |  |  |  |  |  |  |  |  |  |  |  |  |  |  | ✻ |  |  |
| | Pr | ✻ | ✻ |  |  | ✻ | ✻ | ✻ | ✻ |  |  |  |  |  |  |  |  |  |  |  |  |  |  |  |  |  |  | ✻ |  |  |
| | Re | ✻ | ✻ |  |  | ✻ | ✻ | ✻ | ✻ |  |  |  |  |  |  |  |  |  |  |  |  |  |  |  |  |  |  | ✻ |  |  |
| | F1 | ✻ | ✻ |  |  | ✻ | ✻ | ✻ | ✻ |  |  |  |  |  |  |  |  |  |  |  |  |  |  |  |  |  |  | ✻ |  |  |
| | CK | ✻ | ✻ |  |  | ✻ |  |  | ✻ |  |  |  |  |  |  |  |  |  |  |  |  |  |  |  |  |  |  | ✻ |  |  |
| | MCC | ✻ | ✻ |  |  | ✻ | ✻ |  |  |  |  |  |  |  |  |  |  |  |  |  |  |  |  |  |  |  |  | ✻ |  |  |
| | LL | ✻ | ✻ |  |  |  |  |  | ✻ |  |  |  |  |  |  |  |  |  |  |  |  |  |  |  |  |  |  |  |  |  |
| | BAcc | ✻ |  |  |  |  |  |  |  |  |  |  |  |  |  |  |  |  |  |  |  |  |  |  |  |  |  |  |  |  |
| | F-be | ✻ | ✻ |  |  |  |  |  | ✻ |  |  |  |  |  |  |  |  |  |  |  |  |  |  |  |  |  |  | ✻ |  |  |
| | JI | ✻ | ✻ |  |  | ✻ |  |  |  |  |  |  |  |  |  |  |  |  |  |  |  |  |  |  |  |  |  | ✻ |  |  |
| | G-m |  |  |  |  |  |  |  |  | ✻ |  |  |  |  |  |  |  |  |  |  |  |  |  |  |  |  |  |  |  |  |
| | AUC | ✻ | ✻ | ✻ |  |  |  |  | ✻ |  |  |  |  |  |  |  |  |  |  |  |  |  |  |  |  |  |  |  |  |  |
| Regression | MAE | ✻ | ✻ | ✻ |  |  |  |  |  |  |  |  |  |  |  |  |  |  |  |  |  |  |  |  |  |  |  | ✻ |  |  |
| | MSE | ✻ | ✻ | ✻ |  |  |  |  |  |  | ✻ |  |  |  |  |  |  |  |  |  |  |  |  |  |  |  |  | ✻ |  |  |
| | RMSE | ✻ | ✻ | ✻ |  |  |  |  |  |  | ✻ |  |  |  |  |  |  |  |  |  |  |  |  |  |  |  |  | ✻ |  |  |
| | MAPE | ✻ | ✻ | ✻ |  |  |  |  |  |  |  | ✻ | ✻ |  |  |  |  |  |  |  |  |  |  |  |  |  |  | ✻ |  |  |
| | $R^2$ | ✻ |  | ✻ |  |  |  |  |  |  | ✻ | ✻ | ✻ |  |  |  |  |  |  |  |  |  |  |  |  |  |  | ✻ |  |  |
| | MSLE |  | ✻ | ✻ |  |  |  |  |  |  |  |  |  |  |  |  |  |  |  |  |  |  |  |  |  |  |  |  |  |  |
| | EV | ✻ |  | ✻ |  |  |  |  |  |  | ✻ | ✻ | ✻ |  |  |  |  |  |  |  |  |  |  |  |  |  |  | ✻ |  |  |
| | Med AE | ✻ |  |  |  |  |  |  |  |  |  |  |  |  |  |  |  |  |  |  |  |  |  |  |  |  |  |  |  |  |
| | TD | ✻ |  |  |  |  |  |  |  |  |  |  |  |  |  |  |  |  |  |  |  |  |  |  |  |  |  |  |  |  |
| | Hu |  | ✻ |  |  |  |  |  |  |  |  |  |  |  |  |  |  |  |  |  |  |  |  |  |  |  |  |  |  |  |
| Clustering | Sil | ✻ |  |  |  |  |  |  |  |  |  | ✻ |  |  |  |  |  |  |  |  |  |  |  |  |  |  |  |  |  |  |
| | DaBo | ✻ |  |  |  |  |  |  |  |  |  | ✻ |  |  |  |  |  |  |  |  |  |  |  |  |  |  |  |  |  |  |
| | CaHa | ✻ |  |  |  |  |  |  |  |  |  | ✻ |  |  |  |  |  |  |  |  |  |  |  |  |  |  |  |  |  |  |
| | WCSS | ✻ |  |  |  |  |  |  |  |  |  |  |  |  |  |  |  |  |  |  |  |  |  |  |  |  |  |  |  |  |
| Correlation | Pear | ✻ |  | ✻ |  |  |  |  |  |  | ✻ | ✻ | ✻ | ✻ |  | ✻ | ✻ | ✻ | ✻ | ✻ | ✻ | ✻ |  |  |  |  |  |  |  |  |
| | Spea | ✻ |  |  |  |  |  |  |  |  |  | ✻ | ✻ | ✻ | ✻ | ✻ |  | ✻ |  |  |  |  |  |  |  |  |  |  |  |  |
| | Tau |  |  |  |  |  |  |  |  |  |  | ✻ | ✻ | ✻ |  | ✻ |  |  |  |  |  |  |  |  |  |  |  |  |  |  |
| | MI | ✻ |  |  |  |  |  |  |  |  | ✻ |  | ✻ | ✻ |  |  |  |  |  |  |  |  |  |  |  |  |  |  |  |  |
| | DCorr |  |  |  |  |  |  |  |  |  |  |  | ✻ |  | ✻ |  |  |  |  |  |  |  |  |  |  |  |  |  |  |  |
| | Bic |  |  |  |  |  |  |  |  |  |  |  |  |  |  | ✻ |  |  |  |  |  |  |  |  |  |  |  |  |  |  |
| | Perc |  |  |  |  |  |  |  |  |  |  |  |  |  |  | ✻ |  |  |  |  |  |  |  |  |  |  |  |  |  |  |
| | Shep |  |  |  |  |  |  |  |  |  |  |  |  |  |  | ✻ |  |  |  |  |  |  |  |  |  |  |  |  |  |  |
| | PCorr |  |  |  |  |  |  |  |  |  |  |  |  |  |  | ✻ |  |  |  |  |  |  |  |  |  |  |  |  |  |  |
| I2I | MSE | ✻ |  | ✻ |  |  |  |  |  |  | ✻ |  |  |  |  |  |  |  |  |  |  |  |  |  |  |  |  | ✻ |  | ✻ |
| | PSNR |  | ✻ | ✻ |  |  |  |  |  |  |  |  |  |  |  |  |  |  |  |  |  |  |  |  |  |  |  | ✻ |  | ✻ |
| | SSIM |  | ✻ | ✻ |  |  |  |  |  |  |  |  |  |  |  |  |  |  |  |  |  |  |  |  |  |  |  | ✻ |  | ✻ |
| | MAE | ✻ |  | ✻ |  |  |  |  |  |  | ✻ |  |  |  |  |  |  |  |  |  |  |  |  |  |  |  |  | ✻ |  |  |
| | RMSE | ✻ |  | ✻ |  |  |  |  |  |  | ✻ |  |  |  |  |  |  |  |  |  |  |  |  |  |  |  |  | ✻ |  |  |
| | $R^2$ | ✻ |  | ✻ |  |  |  |  |  |  | ✻ |  |  |  |  |  |  |  |  |  |  |  |  |  |  |  |  | ✻ |  |  |
| Segmentation | Acc | ✻ | ✻ | ✻ |  |  |  |  |  |  |  |  |  | ✻ |  |  |  |  |  |  |  |  |  |  |  |  |  | ✻ |  |  |
| | Pr | ✻ | ✻ | ✻ |  |  |  |  |  |  |  |  |  |  |  |  |  |  |  |  |  |  |  |  |  | ✻ |  | ✻ |  |  |
| | Re | ✻ | ✻ | ✻ |  |  |  |  |  |  |  |  |  |  |  |  |  |  |  |  |  |  |  |  |  | ✻ |  | ✻ |  |  |
| | F1 | ✻ | ✻ | ✻ |  |  |  |  |  |  |  |  |  |  |  |  |  |  |  |  |  |  |  |  |  | ✻ |  | ✻ |  |  |
| | IoU | ✻ | ✻ | ✻ |  |  |  |  |  |  |  |  |  | ✻ |  |  |  |  |  |  |  |  |  |  | ✻ | ✻ |  | ✻ |  |  |
| | CK | ✻ |  |  |  |  |  |  |  |  |  |  |  |  |  |  |  |  |  |  |  |  |  |  |  |  |  |  |  |  |
| | AdjR | ✻ |  |  |  |  |  |  |  |  |  |  |  |  |  |  |  |  |  |  |  |  |  |  |  |  |  |  |  |  |
| | M IoU |  | ✻ |  |  |  |  |  |  |  |  |  |  |  |  |  |  |  |  |  |  |  |  |  |  |  |  |  |  |  |
| | BF | ✻ | ✻ | ✻ |  |  |  |  |  |  |  |  |  |  |  |  |  |  |  |  |  |  |  |  | ✻ | ✻ |  | ✻ |  | ✻ |
| | Haus |  | ✻ | ✻ |  |  |  |  |  |  |  |  |  | ✻ |  |  |  |  |  |  |  |  |  |  | ✻ | ✻ | ✻ | ✻ | ✻ | ✻ |
| | Dice |  |  | ✻ |  |  |  |  |  |  |  |  |  | ✻ |  |  |  |  |  |  |  |  |  |  | ✻ | ✻ |  | ✻ |  |  |
| | AdaR |  |  |  |  |  |  |  |  |  |  |  |  |  |  |  |  |  |  |  |  |  |  |  |  |  |  |  |  | ✻ |
| | VoI |  |  |  |  |  |  |  |  |  |  |  |  |  |  |  |  |  |  |  |  |  |  |  |  |  |  |  |  | ✻ |
| | G-m |  |  |  |  |  |  |  |  | ✻ |  |  |  |  |  |  |  |  |  |  |  |  |  |  |  |  |  |  |  |  |
| Statistical Tests | itt |  |  |  |  |  |  |  |  |  | ✻ | ✻ |  |  |  | ✻ |  |  |  |  |  |  | ✻ | ✻ |  |  |  |  |  |  |
| | ptt |  |  |  |  |  |  |  |  |  | ✻ | ✻ |  |  |  | ✻ |  |  |  |  |  |  | ✻ | ✻ |  |  |  |  |  |  |
| | kst |  |  |  |  |  |  |  |  |  | ✻ | ✻ |  |  |  |  |  |  |  |  |  |  |  |  |  |  |  |  |  |  |
| | chi2 |  |  |  |  |  |  |  |  |  |  | ✻ |  |  |  |  |  |  |  |  |  |  |  |  |  |  |  |  |  |  |
| | ANO |  |  |  |  |  |  |  |  |  | ✻ | ✻ |  |  |  | ✻ |  |  |  |  |  |  |  |  |  |  |  |  |  |  |
| | kwt |  |  |  |  |  |  |  |  |  |  | ✻ |  |  |  | ✻ |  |  |  |  |  |  |  |  |  |  |  |  |  |  |
| | mwut |  |  |  |  |  |  |  |  |  |  | ✻ |  |  |  | ✻ |  |  |  |  |  |  |  |  |  |  |  |  |  |  |
| | swt |  |  |  |  |  |  |  |  |  |  | ✻ |  |  |  | ✻ |  |  |  |  |  |  |  |  |  |  |  |  |  |  |
| | ft |  |  |  |  |  |  |  |  |  |  | ✻ |  |  |  |  |  |  |  |  |  |  |  |  |  |  |  |  |  |  |
| | pt |  |  |  |  |  |  |  |  |  |  | ✻ |  |  |  |  |  |  |  |  |  |  |  |  |  |  |  |  |  |  |
| | wet |  |  |  |  |  |  |  |  |  | ✻ | ✻ |  |  |  | ✻ |  |  |  |  |  |  |  | ✻ |  |  |  |  |  |  |
| | bt |  |  |  |  |  |  |  |  |  |  | ✻ |  |  |  |  |  |  |  |  |  |  |  |  |  |  |  |  |  |  |
| | lt |  |  |  |  |  |  |  |  |  |  | ✻ |  |  |  |  |  |  |  |  |  |  |  |  |  |  |  |  |  |  |
| | wsrt |  |  |  |  |  |  |  |  |  |  | ✻ |  |  |  |  |  |  |  |  |  |  |  | ✻ |  |  |  |  |  |  |
| | zt |  |  |  |  |  |  |  |  |  | ✻ |  |  |  |  | ✻ |  |  |  |  |  |  |  |  |  |  |  |  |  |  |

| | | R | | | | | | | | | | | | | | | | | | | | | | | | | | | | | | | Matlab |
|---|---|---|---|---|---|---|---|---|---|---|---|---|---|---|---|---|---|---|---|---|---|---|---|---|---|---|---|---|---|---|---|---|---|
| Tasks | Metrics | Caret | MLmetrics | Yardstick | Psych | E1071 | IRR | MCCR | Cluster | Factoextra | ClusterCrit | Infotheo | Energy | Entropy | Corrplot | Hmisc | Metrics | ModelMetrics | DescTools | Imager | Forecast | hydroGOF | TDR | Rminer | MetricsWeighted | Mclust | Infer | Rstatix | Car | Stats | Base R | pROC | ROCR | Matlab |
| Classification | Acc | ✻ | ✻ | ✻ |  |  |  |  |  |  |  |  |  |  |  |  |  |  |  |  |  |  |  |  |  |  |  |  |  |  | ✻ |  |  | ✻ |
| | Pr | ✻ | ✻ | ✻ |  |  |  |  |  |  |  |  |  |  |  |  |  |  |  |  |  |  |  |  |  |  |  |  |  |  | ✻ |  |  | ✻ |
| | Re | ✻ | ✻ | ✻ |  |  |  |  |  |  |  |  |  |  |  |  |  |  |  |  |  |  |  |  |  |  |  |  |  |  | ✻ |  |  | ✻ |
| | F1 | ✻ | ✻ | ✻ |  |  |  |  |  |  |  |  |  |  |  |  |  |  |  |  |  |  |  |  |  |  |  |  |  |  | ✻ |  |  | ✻ |
| | CK | ✻ |  | ✻ | ✻ | ✻ | ✻ |  |  |  |  |  |  |  |  |  |  |  |  |  |  |  |  |  |  |  |  |  |  |  | ✻ |  |  | ✻ |
| | MCC | ✻ |  |  |  |  |  | ✻ |  |  |  |  |  |  |  |  |  |  |  |  |  |  |  |  |  |  |  |  |  |  | ✻ |  |  | ✻ |
| | LL |  |  |  |  |  |  |  |  |  |  |  |  |  |  |  |  |  |  |  |  |  |  |  |  |  |  |  |  |  |  |  |  | ✻ |
| | BAcc | ✻ |  | ✻ |  |  |  |  |  |  |  |  |  |  |  |  |  |  |  |  |  |  |  |  |  |  |  |  |  |  | ✻ |  |  | ✻ |
| | F-be |  |  | ✻ |  |  |  |  |  |  |  |  |  |  |  |  |  |  |  |  |  |  |  |  |  |  |  |  |  |  | ✻ |  |  | ✻ |
| | JI | ✻ |  |  |  |  |  |  |  |  |  |  |  |  |  |  |  |  |  |  |  |  |  |  |  |  |  | ✻ |  |  | ✻ |  |  | ✻ |
| | G-m | ✻ |  | ✻ |  |  |  |  |  |  |  |  |  |  |  |  |  |  |  |  |  |  |  |  |  |  |  |  |  |  | ✻ |  |  | ✻ |
| | AUC |  |  |  |  |  |  |  |  |  |  |  |  |  |  |  |  |  |  |  |  |  |  |  |  |  |  |  |  |  |  | ✻ | ✻ | ✻ |
| Regression | MAE | ✻ | ✻ | ✻ |  |  |  |  |  |  |  |  |  |  |  |  | ✻ | ✻ |  |  | ✻ | ✻ | ✻ | ✻ |  |  |  |  |  |  | ✻ |  |  | ✻ |
| | MSE |  | ✻ |  |  |  |  |  |  |  |  |  |  |  |  |  | ✻ | ✻ |  |  | ✻ | ✻ | ✻ | ✻ |  |  |  |  |  |  | ✻ |  |  | ✻ |
| | RMSE | ✻ | ✻ | ✻ |  |  |  |  |  |  |  |  |  |  |  |  | ✻ | ✻ |  |  | ✻ | ✻ | ✻ | ✻ |  |  |  |  |  |  | ✻ |  |  | ✻ |
| | MAPE |  | ✻ | ✻ |  |  |  |  |  |  |  |  |  |  |  |  | ✻ |  |  |  |  |  |  |  |  |  |  |  |  |  | ✻ |  |  | ✻ |
| | $R^2$ | ✻ | ✻ | ✻ |  |  |  |  |  |  |  |  |  |  |  |  |  |  |  |  |  |  | ✻ | ✻ |  |  |  |  |  |  | ✻ |  |  | ✻ |
| | MSLE |  |  |  |  |  |  |  |  |  |  |  |  |  |  |  | ✻ | ✻ |  |  |  |  |  |  |  |  |  |  |  |  | ✻ |  |  | ✻ |
| | EV |  |  |  |  |  |  |  |  |  |  |  |  |  |  |  |  |  |  |  |  |  |  |  |  |  |  |  |  |  | ✻ |  |  | ✻ |
| | Med AE |  | ✻ |  |  |  |  |  |  |  |  |  |  |  |  |  | ✻ |  |  |  |  |  |  |  |  |  |  |  |  |  | ✻ |  |  | ✻ |
| | TD |  |  |  |  |  |  |  |  |  |  |  |  |  |  |  |  |  |  |  |  |  |  |  | ✻ |  |  |  |  |  |  |  |  | ✻ |
| | Hu |  |  | ✻ |  |  |  |  |  |  |  |  |  |  |  |  |  |  |  |  |  |  |  |  |  |  |  |  |  |  | ✻ |  |  | ✻ |
| Clustering | Sil |  |  |  |  |  |  |  | ✻ | ✻ |  |  |  |  |  |  |  |  |  |  |  |  |  |  |  |  |  |  |  |  |  |  |  | ✻ |
| | DaBo |  |  |  |  |  |  |  |  |  | ✻ |  |  |  |  |  |  |  |  |  |  |  |  |  |  |  |  |  |  |  |  |  |  | ✻ |
| | CaHa |  |  |  |  |  |  |  |  |  | ✻ |  |  |  |  |  |  |  |  |  |  |  |  |  |  |  |  |  |  |  |  |  |  | ✻ |
| | WCSS |  |  |  |  |  |  |  |  |  |  |  |  |  |  |  |  |  |  |  |  |  |  |  |  |  |  |  |  | ✻ |  |  |  | ✻ |
| Correlation | Pear |  |  |  | ✻ |  |  |  |  |  |  |  |  |  | ✻ | ✻ |  |  |  |  |  |  |  |  |  |  |  |  |  |  | ✻ |  |  | ✻ |
| | Spea |  |  |  | ✻ |  |  |  |  |  |  |  |  |  | ✻ | ✻ |  |  |  |  |  |  |  |  |  |  |  |  |  |  | ✻ |  |  | ✻ |
| | Tau |  |  |  | ✻ |  |  |  |  |  |  |  |  |  | ✻ |  |  |  |  |  |  |  |  |  |  |  |  |  |  |  | ✻ |  |  | ✻ |
| | MI |  |  |  |  |  |  |  |  |  |  | ✻ |  | ✻ |  |  |  |  |  |  |  |  |  |  |  |  |  |  |  |  |  |  |  | ✻ |
| | DCorr |  |  |  |  |  |  |  |  |  |  |  | ✻ |  |  |  |  |  |  |  |  |  |  |  |  |  |  |  |  |  |  |  |  | ✻ |
| | Bic |  |  |  |  |  |  |  |  |  |  |  |  |  |  |  |  |  |  |  |  |  |  |  |  |  |  |  |  |  |  |  |  | ✻ |
| | Perc |  |  |  |  |  |  |  |  |  |  |  |  |  |  |  |  |  |  |  |  |  |  |  |  |  |  |  |  |  |  |  |  |  |
| | Shep |  |  |  |  |  |  |  |  |  |  |  |  |  |  |  |  |  |  |  |  |  |  |  |  |  |  |  |  |  |  |  |  |  |
| | PCorr |  |  |  |  |  |  |  |  |  |  |  |  |  |  |  |  |  |  |  |  |  |  |  |  |  |  |  |  |  |  |  |  |  |
| I2I | MSE |  | ✻ |  |  |  |  |  |  |  |  |  |  |  |  |  | ✻ | ✻ | ✻ |  |  |  |  |  |  |  |  |  |  |  | ✻ |  |  | ✻ |
| | PSNR |  |  |  |  |  |  |  |  |  |  |  |  |  |  |  |  |  |  | ✻ |  |  |  |  |  |  |  |  |  |  | ✻ |  |  | ✻ |
| | SSIM |  |  |  |  |  |  |  |  |  |  |  |  |  |  |  |  |  |  |  |  |  |  |  |  |  |  |  |  |  | ✻ |  |  | ✻ |
| | MAE | ✻ | ✻ | ✻ |  |  |  |  |  |  |  |  |  |  |  |  | ✻ | ✻ | ✻ |  |  |  |  |  |  |  |  |  |  |  | ✻ |  |  | ✻ |
| | RMSE | ✻ |  | ✻ |  |  |  |  |  |  |  |  |  |  |  |  | ✻ | ✻ | ✻ |  |  |  |  |  |  |  |  |  |  |  | ✻ |  |  | ✻ |
| | $R^2$ | ✻ | ✻ | ✻ |  |  |  |  |  |  |  |  |  |  |  |  |  |  | ✻ |  |  |  |  |  |  |  |  |  |  |  | ✻ |  |  | ✻ |
| Segmentation | Acc |  | ✻ | ✻ |  |  |  |  |  |  |  |  |  |  |  |  | ✻ |  |  |  |  |  |  |  |  |  |  |  |  |  | ✻ |  |  | ✻ |
| | Pr |  | ✻ | ✻ |  |  |  |  |  |  |  |  |  |  |  |  | ✻ |  |  |  |  |  |  |  |  |  |  |  |  |  | ✻ |  |  | ✻ |
| | Re |  | ✻ |  |  |  |  |  |  |  |  |  |  |  |  |  | ✻ | ✻ |  |  |  |  |  |  |  |  |  |  |  |  | ✻ |  |  | ✻ |
| | F1 |  | ✻ |  |  |  |  |  |  |  |  |  |  |  |  |  |  |  |  |  |  |  |  |  |  |  |  |  |  |  |  |  |  |  |
| | IoU |  |  |  |  |  |  |  |  |  |  |  |  |  |  |  |  |  |  |  |  |  |  |  |  |  |  |  |  |  | ✻ |  |  | ✻ |
| | CK |  |  | ✻ |  |  |  |  |  |  |  |  |  |  |  |  |  |  |  |  |  |  |  |  |  |  |  |  |  |  | ✻ |  |  | ✻ |
| | AdjR |  |  |  |  |  |  |  |  |  |  |  |  |  |  |  |  |  |  |  |  |  |  |  |  | ✻ |  |  |  |  | ✻ |  |  | ✻ |
| | M IoU |  |  |  |  |  |  |  |  |  |  |  |  |  |  |  |  |  |  |  |  |  |  |  |  |  |  |  |  |  | ✻ |  |  | ✻ |
| | BF |  |  |  |  |  |  |  |  |  |  |  |  |  |  |  |  |  |  |  |  |  |  |  |  |  |  |  |  |  | ✻ |  |  | ✻ |
| | Haus |  |  |  |  |  |  |  |  |  |  |  |  |  |  |  |  |  |  |  |  |  |  |  |  |  |  |  |  |  | ✻ |  |  | ✻ |
| | Dice |  |  |  |  |  |  |  |  |  |  |  |  |  |  |  |  |  |  |  |  |  |  |  |  |  |  |  |  |  | ✻ |  |  | ✻ |
| | AdaR |  |  |  |  |  |  |  |  |  |  |  |  |  |  |  |  |  |  |  |  |  |  |  |  |  |  |  |  |  | ✻ |  |  | ✻ |
| | VoI |  |  |  |  |  |  |  |  |  |  |  |  |  |  |  |  |  |  |  |  |  |  |  |  |  |  |  |  |  | ✻ |  |  |  |
| | G-m | ✻ |  | ✻ |  |  |  |  |  |  |  |  |  |  |  |  |  |  |  |  |  |  |  |  |  |  |  |  |  |  | ✻ |  |  | ✻ |
| Statistical Tests | itt |  |  |  |  |  |  |  |  |  |  |  |  |  |  |  |  |  |  |  |  |  |  |  |  |  |  |  |  |  | ✻ |  |  | ✻ |
| | ptt |  |  |  |  |  |  |  |  |  |  |  |  |  |  |  |  |  |  |  |  |  |  |  |  |  |  |  |  | ✻ | ✻ |  |  | ✻ |
| | kst |  |  |  |  |  |  |  |  |  |  |  |  |  |  |  |  |  |  |  |  |  |  |  |  |  |  |  |  |  | ✻ |  |  | ✻ |
| | chi2 |  |  |  |  |  |  |  |  |  |  |  |  |  |  |  |  |  |  |  |  |  |  |  |  |  |  |  |  |  | ✻ |  |  | ✻ |
| | ANO |  |  |  |  |  |  |  |  |  |  |  |  |  |  |  |  |  |  |  |  |  |  |  |  |  |  |  | ✻ |  | ✻ |  |  | ✻ |
| | kwt |  |  |  |  |  |  |  |  |  |  |  |  |  |  |  |  |  |  |  |  |  |  |  |  |  |  | ✻ |  |  | ✻ |  |  |  |
| | mwut |  |  |  |  |  |  |  |  |  |  |  |  |  |  |  |  |  |  |  |  |  |  |  |  |  |  |  |  |  | ✻ |  |  | ✻ |
| | swt |  |  |  |  |  |  |  |  |  |  |  |  |  |  |  |  |  |  |  |  |  |  |  |  |  |  |  |  |  | ✻ |  |  | ✻ |
| | ft |  |  |  |  |  |  |  |  |  |  |  |  |  |  |  |  |  |  |  |  |  |  |  |  |  |  |  |  |  | ✻ |  |  | ✻ |
| | pt |  |  |  |  |  |  |  |  |  |  |  |  |  |  |  |  |  |  |  |  |  |  |  |  |  | ✻ |  |  |  | ✻ |  |  |  |
| | wet |  |  |  |  |  |  |  |  |  |  |  |  |  |  |  |  |  |  |  |  |  |  |  |  |  |  |  |  |  | ✻ |  |  | ✻ |
| | bt |  |  |  |  |  |  |  |  |  |  |  |  |  |  |  |  |  |  |  |  |  |  |  |  |  |  |  |  |  | ✻ |  |  | ✻ |
| | lt |  |  |  |  |  |  |  |  |  |  |  |  |  |  |  |  |  |  |  |  |  |  |  |  |  |  |  |  |  | ✻ |  |  | ✻ |
| | wsrt |  |  |  |  |  |  |  |  |  |  |  |  |  |  |  |  |  |  |  |  |  |  |  |  |  |  |  |  |  | ✻ |  |  | ✻ |
| | zt |  |  |  |  |  |  |  |  |  |  |  |  |  |  |  |  |  |  |  |  |  |  |  |  |  |  |  |  | ✻ | ✻ |  |  | ✻ |

## 3. RESULTS

In this evaluation study, we focused on the differences between the values of evaluation metrics calculated by various components across Python, R, and Matlab programming languages, rounded to two decimal points. Figure 1 displays only some metrics that indicate inconsistencies across different components in various programming languages, while Supplemental Figure S1-S30 presents all values calculated by different components, and various datasets including those with consistent values.

### 3.1 Results for Classification Tasks:

**Binary Classification Tasks**. In binary classification, we employed D1 as input features and binary overall survival as a true outcome. For Precision metric, most components yielded a value of 0.09 for class 1. However, PyCM in Python reported an average of 0.44, with 0.78 for class 0 and 0.09 for class 1. Similarly, Caret and Yardstick in R recorded 0.78 for class 0, as depicted in Figure 1 (i/a). Regarding Recall, PyCM reported an average of 0.46, with 0.86 for class 0 and 0.05 for class 1. Additionally, Caret and Yardstick calculated a value of 0.86 for class 0, as shown in Figure 1 (i/b). When evaluating F1 Score, PyCM provided an average of 0.44, with 0.81 for class 0 and 0.07 for class 1, aligning with Caret and Yardstick for class 0, while other components reported 0.07 for class 1, as illustrated in Figure 1 (i/c). For the Jaccard Index, PyCM reported an average of 0.36, with 0.69 for class 0 and 0.04 for class 1. TensorFlow in Python averaged the values across both classes, resulting in 0.36, while other components reported 0.04 for class 1, as depicted in Figure 1 (i/d). Other metrics across various components showed similar values, as detailed in Supplemental Figure S1. The results were further validated using additional datasets, including D2 and D3 (with overall survival as the true outcome), D8 (with Gleason Score as the true outcome) and randomly selected dataset. These validations are presented in Supplemental Figures S2-S5.

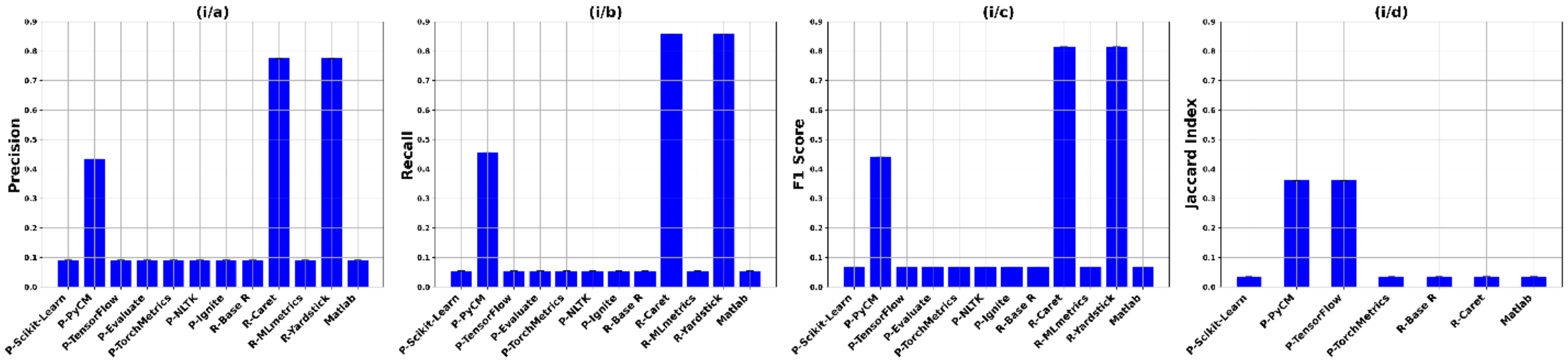

**Figure 1.** Inconsistent evaluation metrics in binary classification.

**Multi-class Classification Tasks.** In multi-class classification, we employed D8 as input features and Gleason scores as a true outcome. For Precision, all components returned values of 0.94, 0.60, and 0.90 for the three classes, respectively. However, TensorFlow and TorchMetrics in Python reported a micro Precision of 0.79, as depicted in Figure 2 (ii/a). Similarly, for Recall, all components reported values of 0.90, 0.88, and 0.53 for the three classes, while TensorFlow and TorchMetrics provided a micro Recall of 0.79, as seen in Figure 2 (ii/b). For F1 Score, the most components reported 0.92, 0.71, and 0.67 for the three classes, while TorchMetrics reported a micro F1 Score of 0.79, as depicted in Figure 2 (ii/c). Regarding MCC, PyCM reported values of 0.86, 0.59, and 0.61 for the three classes, with an average of 0.68, while other components reported a fixed value of 0.69, as shown in Figure 2 (ii/d). For Balanced Accuracy, Matlab and Scikit-Learn in Python provided a weighted accuracy of 0.77, Base R and Caret in R reported 0.79, and Yardstick showed a macro Accuracy of 0.83 across the three classes, as depicted in Figure 2 (ii/e). Lastly, for Jaccard Index, most components returned values of 0.85, 0.56, and 0.50, while TensorFlow reported 0.73 and TorchMetrics returned 0.63, as depicted in Figure 2 (ii/f). Other metrics across various components showed similar values, as detailed in Supplemental Figure S6. The results were further validated using a randomly selected dataset, as presented in Supplemental Figure S7.

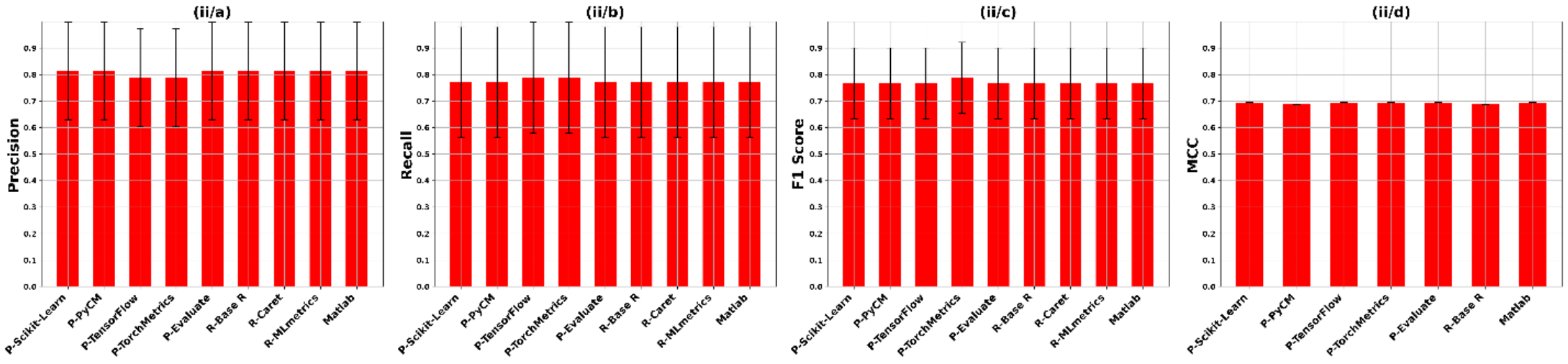

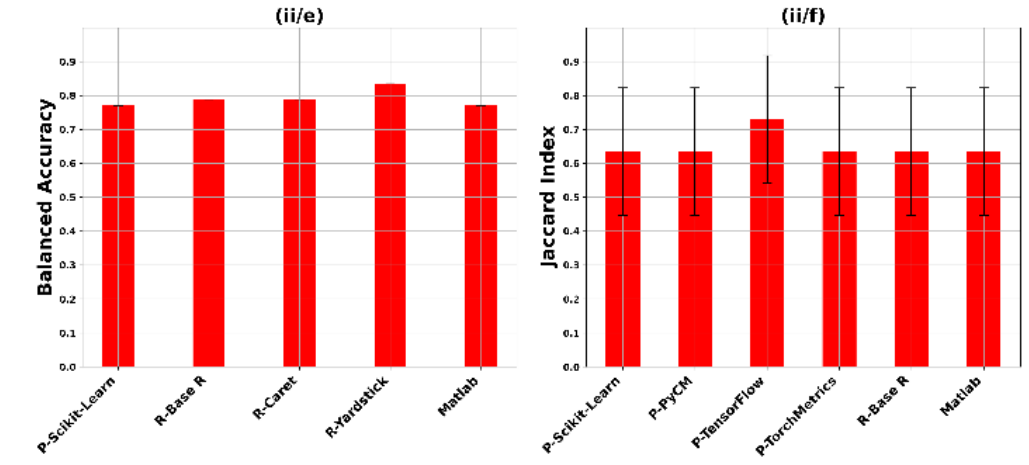

**Figure 2.** Inconsistent evaluation metrics in multi-class classification
MCC: Matthew's Correlation Coefficient.

### 3.2 Results for Regression Tasks:

This task employed D1 as input features and continuous overall survival time as a true outcome. For R², Scikit-Learn and TorchMetrics in Python, Base R and MLmetrics in R reported a value of -0.03, while StatsModels, SciPy, and Pandas in Python, and Yardstick, Caret, TDR, and Rminer in R, and Matlab returned a value of 0.02, as shown in Figure 3 (iii/a). For Tweedie Deviance, Matlab and Scikit-Learn both reported 749,309.73, whereas MetricsWeighted package in R returned 749,549.57 for power = 0, as plotted in Figure 3 (iii/b). Other metrics across various components showed similar values, as detailed in Supplemental Figure S8. The results were further validated with additional datasets, including D2 and D3 (continuous overall survival time as the outcome) and random selected dataset, as shown in Supplemental Figures S9-S11.

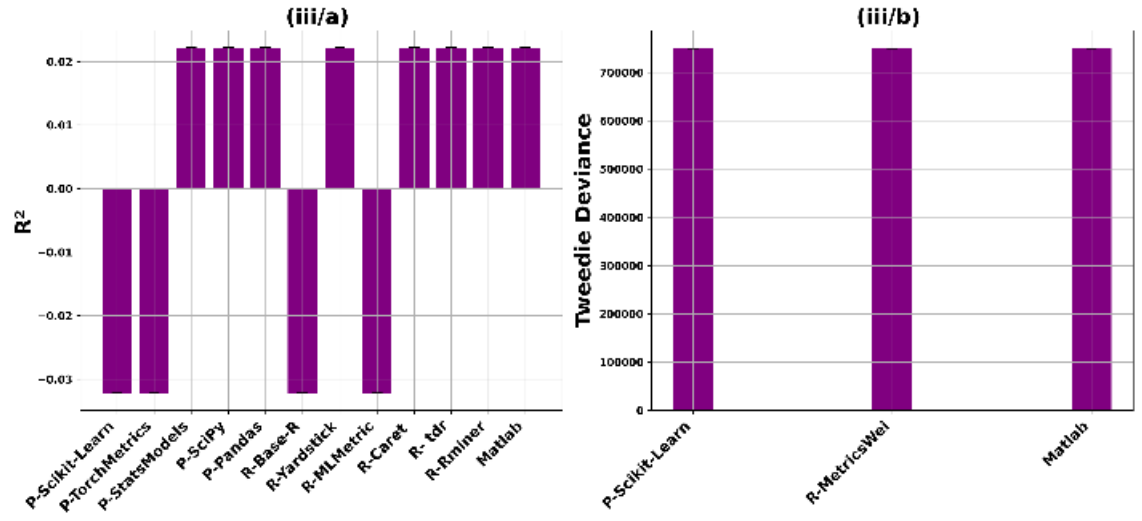

**Figure 3.** Inconsistent evaluation metrics in regression
R²: R-Squared.

### 3.3 Results for Clustering Tasks:

This task utilized D4 as input features. For Silhouette Score, SciPy in Python gave 0.11, and the other libraries returned a value of 0.09, as illustrated in Figure 4 (iv/a). For WCSS, Scikit-Learn in Python provided a value of 92,307.43, Stats in R returned 91,190.25, and Matlab reported 1,199,484.88, as demonstrated in Figure 4 (iv/b). Other metrics remain consistent to two decimal places, as detailed in Supplemental Figure S12. The results were further validated with additional datasets, including D1, D2, D3, D5, D6, D7 and a random selected dataset, as shown in Supplemental Figures S13-S19.

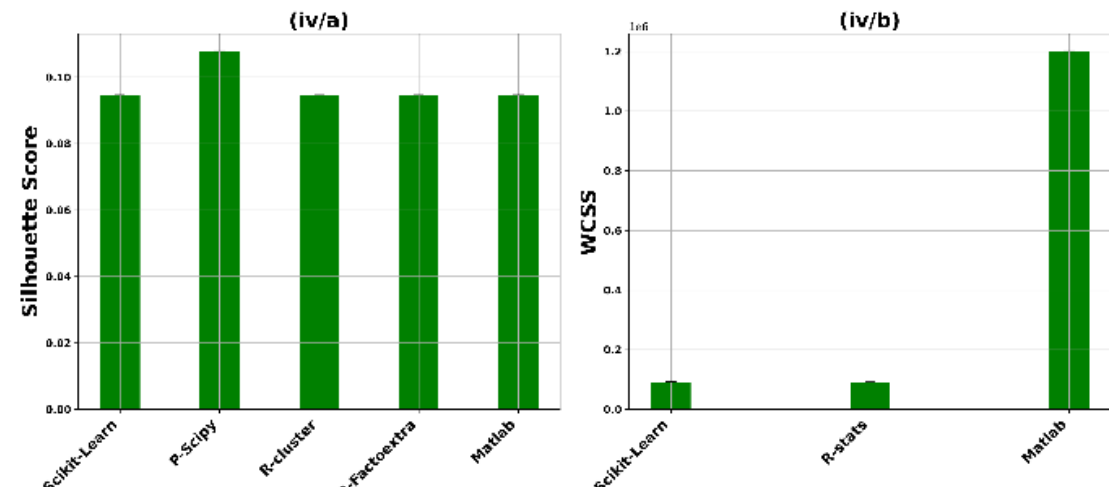

**Figure 4.** Inconsistent evaluation metrics in clustering
WCSS: Within-Cluster Sum of Squares.

### 3.4 Results for Correlation Tasks:

This task utilized D9 as input features. For the Bicor metric, Pingouin in Python provided a value of 0.91, while Matlab gave 0.92, as shown in Figure 5 (v/a). The other metrics are consistent, as detailed in Supplemental Figure S20. The results were further assessed with additional datasets, including D10, D11, and a random selected dataset, as shown in Supplemental Figures S21-S23.

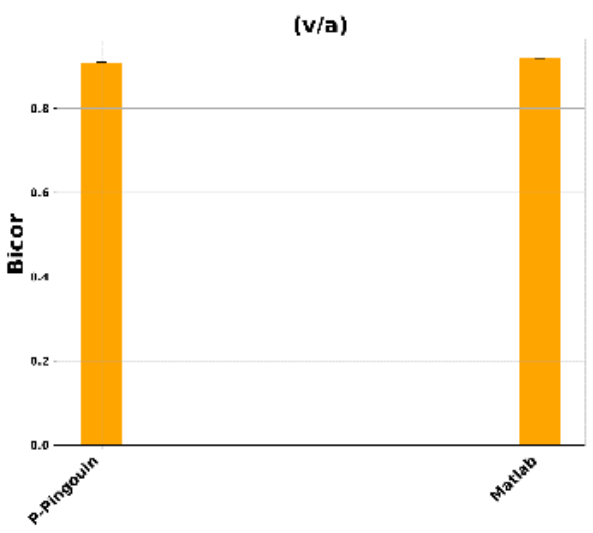

**Figure 5.** Inconsistent evaluation metric in correlation analysis

### 3.5 Results for Statistical Tasks:

This task utilized a random selected dataset as input features. For p-value of independent t-test, all components returned 0.49, but SymPy in Python reported 1.51, as shown in Figure 6 (vi/a). For the p-value of Kolmogorov-Smirnov Test, all components returned 0.39, but Matlab reported 0.38, as depicted in Figure 6 (vi/b). For statistical Mann-Whitney U Test, all components returned 19,228.00, but Matlab reported 40,872.00, as illustrated in Figure 6 (vi/c). For p-value of the F-test, all components returned 0.49, but Matlab reported 0.93, as presented in Figure 6 (vi/d). For p-value of Permutation test, SciPy returned 0.49, Base R reported 0.48, and Infer in R returned 0.50, as shown in Figure 6 (vi/e). For statistical Levene's test, all components returned 0.23, but Matlab reported 0.18, as depicted in Figure 6 (vi/f). For p-value of Levene's test, all components returned 0.63, but Matlab reported 0.67, as shown in Figure 6 (vi/g). Other statistical tests remain consistent across the components, as detailed in Supplemental Figure S24. The results were further assessed with additional datasets, including D9, D10, and D11, as shown in Supplemental Figures S25-S27.

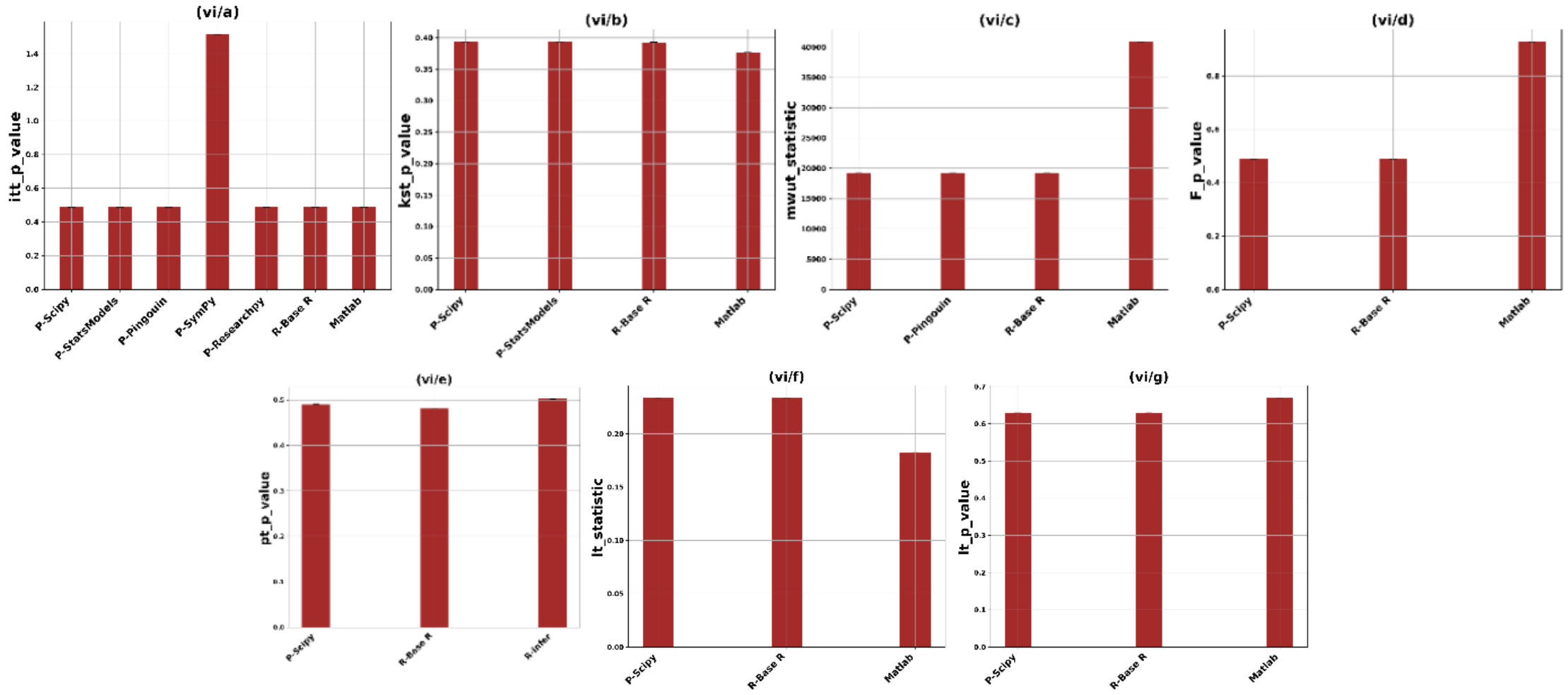

**Figure 6.** Inconsistent evaluation metric in statistical tests

itt: independent t-test; kst: Kolmogorov-Smirnov Test; mwut: Mann-Whitney U Test; pt: Permutation test; lt: Levene's test.

### 3.6 Results for Segmentation Tasks:

**2D-Segmentation Tasks.** This task utilized D12 including masks generated by 2D MEDSAM segmentation network as input images. For IoU, all components returned a value of 0.003±0.01, while TensorFlow in Python returned a value of 0.50±0.005, as shown in Figure 7 (vii/a). For Hausdorff Distance, TochMetrics and MedPy in Python reported 322.15, Scikit-Image in Python returned 309.72 and Monai in Python gave 267.67, as depicted in Figure 7 (vii/b). For Dice Coefficient, all components returned a value of 0.007±0.02, while TorchMetrics gave 0.99±0.005, as shown in Figure 7 (vii/c). Other metrics remain consistent across the components, as detailed in Supplemental Figure 28. The results were further validated with additional two random-selected datasets, as shown in Supplemental Figures S29-S30.

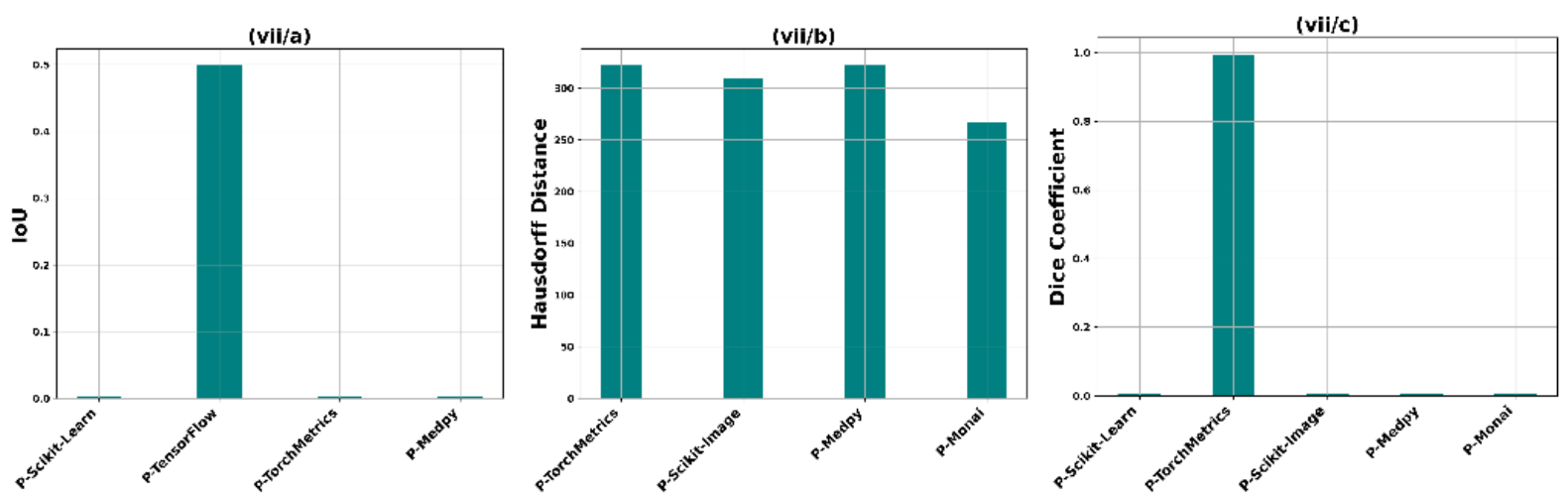

**Figure 7.** Inconsistent evaluation metrics in 2D-segmentation
IoU: Intersection over Union.

**3D-Segmentation Tasks.** This task utilized D13 including masks generated by 3D ReconNet segmentation networks as input images. For Precision, Scikit-Learn in Python returned 0.54±0.23, MedPy in Python gave 0.53±0.22, TensorFlow in Python reported 0.44±0.29, while TorchMetrics in Python returned 0.63±0.38, as shown in Figure 8 (viii/a). For Recall, Scikit-Learn returned 0.78±0.31, MedPy gave 0.79±0.30, TensorFlow reported 0.66±0.41, while TorchMetrics returned 0.36±0.30, as depicted in Figure 8 (viii/b). For F1 Score, Scikit-Learn and MedPy returned 0.58±0.23, while TorchMetrics returned 0.43±0.31, as shown in Figure 8 (viii/c). For IoU, Scikit-Learn and MedPy returned 0.45±0.21, TensorFlow reported 0.71±0.14, while TorchMetrics returned 0.32±0.26, as shown in Figure 8 (viii/d). For BF Score, Scikit-Learn returned 0.58±0.23, TensorFlow reported 0.46±0.29, while TorchMetrics returned 0.59±0.34, as represented in Figure 8 (viii/e). For Hausdorff Distance, TensorFlow returned 17.82, TorchMetrics reported 38.15, while MedPy returned 144501.48, as shown in Figure 8 (viii/f). For Dice Coefficient, TorchMetrics gave 0.99 while MedPy and SimpleITK in Python returned 0.58±0.23, as depicted in Figure 8 (viii/g). Other metrics remain consistent across the components, as detailed in Supplemental Figure S31. The results were further validated with an additional random selected dataset, as shown in Supplemental Figure S32.

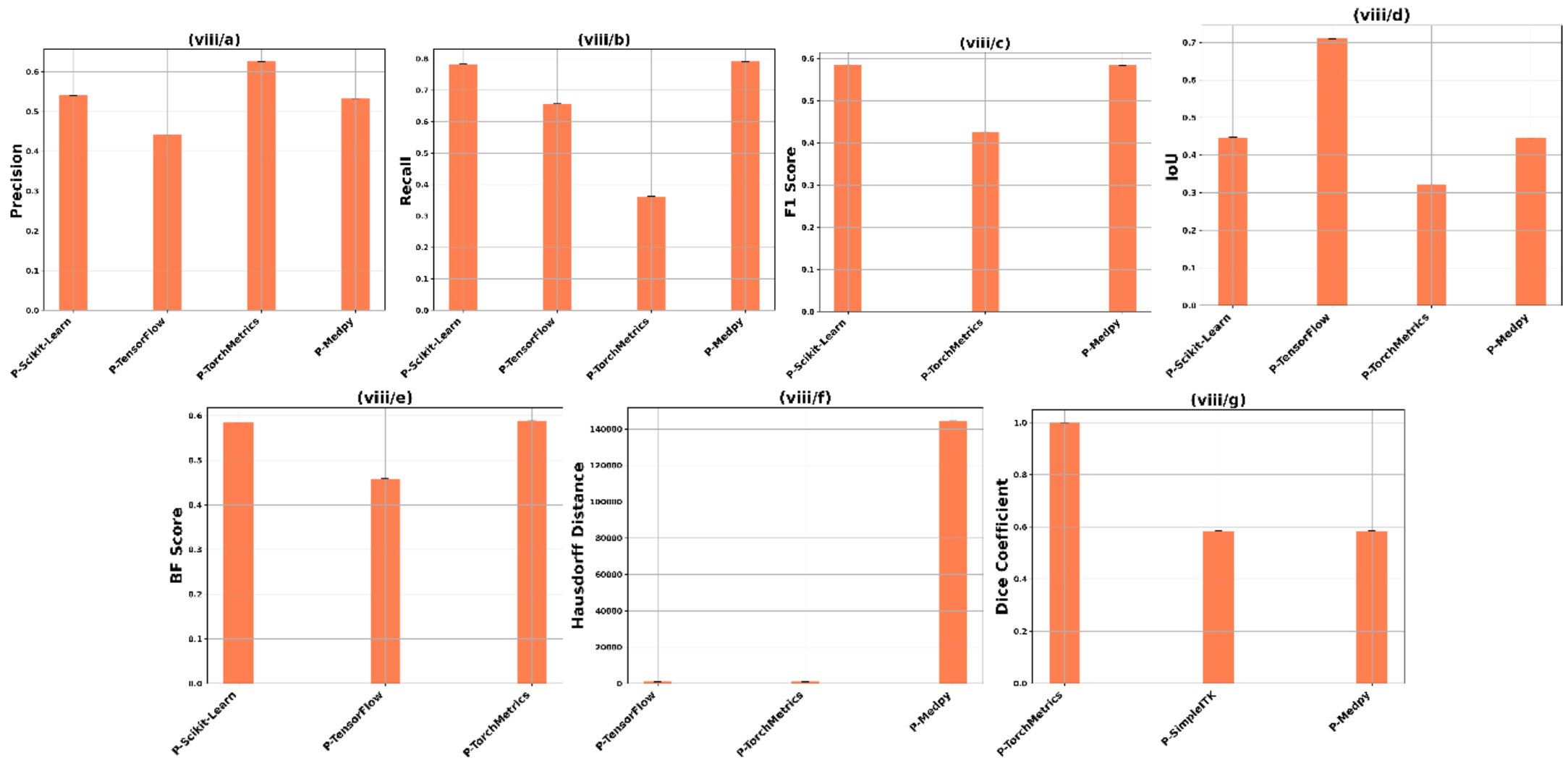

**Figure 8.** Inconsistent evaluation metrics in 3D-segmentation
IoU: Intersection over Union; BF Score: Boundary F1 Score.

### 3.7 Results for I2I Translation Tasks:

**2D-I2I Translation Tasks.** This task utilized D14 including synthetic MRI images generated by 2D CycleGAN networks as input images. For PSNR, Scikit-Image in Python returned -42.96, while TensorFlow and TorchMetrics in Python resulted in -42.97, as shown in Figure 9 (ix/a). For SSIM, Scikit-Image and TensorFlow resulted the same value of 0.0001, while TorchMetrics gave 0.10, as depicted in Figure 9 (ix/b). Other metrics remain consistent across the components, as detailed in Supplemental Figure S33. The results were further validated with an additional random selected dataset, as shown in Supplemental Figure S34.

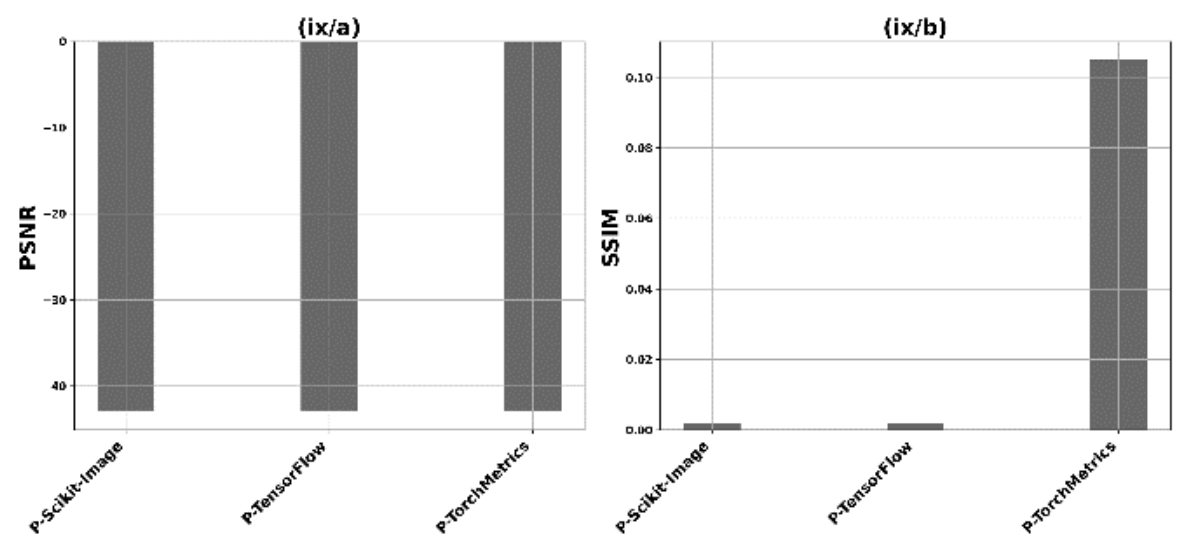

**Figure 9.** Inconsistent evaluation metrics in 2D-I2I translation
PSNR: Peak Signal-to-Noise Ratio; and SSIM: Structural Similarity Index Measure.

**3D-I2I Translation Tasks.** This task utilized D15 including synthetic MRI images generated by 3D CycleGAN networks as input images. For PSNR, Scikit-Image returned 18.09, TensorFlow in Python returned 12.12, while TorchMetrics reported 6.81, as shown in Figure 10 (x/a). For SSIM, Scikit-Image gave 0.10, while TensorFlow and TorchMetrics returned 0.16 as represented in Figure 10 (x/b). For $R^2$, Scikit-Learn and Statsmodels in Python returned -1.09 and 0.002, respectively, as shown in Figure 10 (x/c). The other metrics remain consistent across the components, as detailed in Supplemental Figure S35. The results were further validated with an additional random selected dataset, as shown in Supplemental Figure S36.

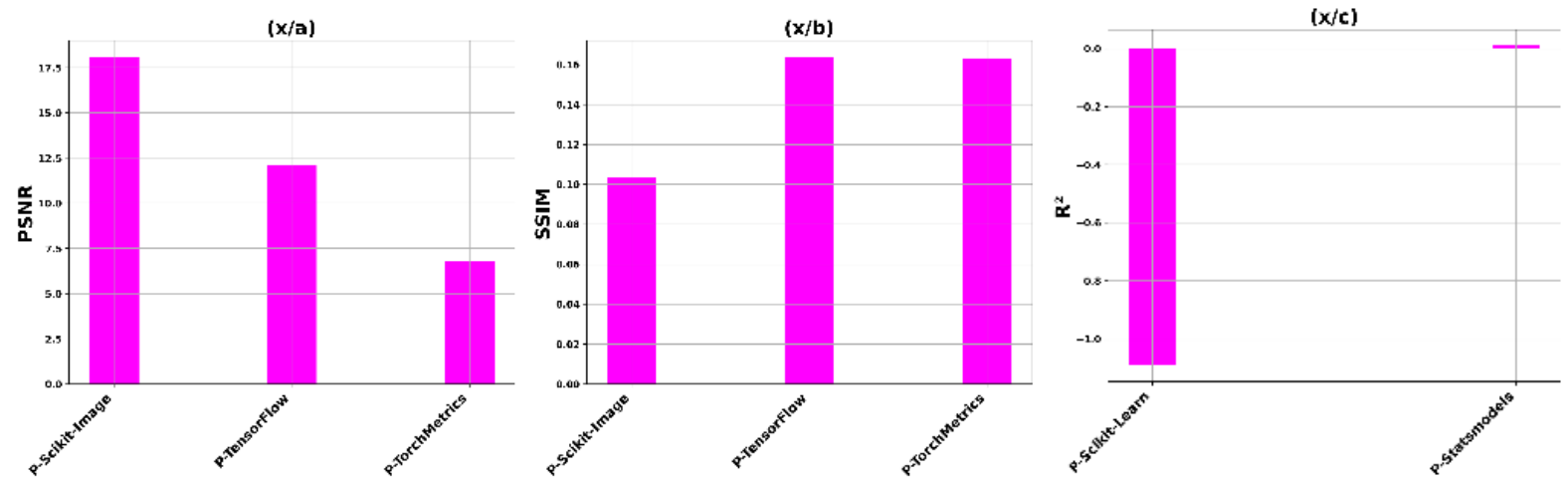

**Figure 10.** Inconsistent evaluation metrics in 3D-I2I translation
PSNR: Peak Signal-to-Noise Ratio; and SSIM: Structural Similarity Index Measure; $R^2$: R-Squared.

## 4. DISCUSSION

Over the past decades, computational analysis has become increasingly widespread across different fields, driving rapid advancements in ML [1, 2]. As ML applications expand into areas like healthcare, finance, engineering, and transporation, ensuring consistency in evaluation is crucial for reliable and comparable performance assessments. Variations in programming languages and novel methods can cause discrepancies in evaluations, leading to inconsistent evaluation of a model's true performance and complicating identification of effective solutions. An important challenge in ML evaluation, which we aimed to evaluate, is the variety and inconsistency in how these metrics are implemented across different programming languages.

Standardized evaluation practices minimize these differences, enabling fair benchmarking, accurate comparisons, and meaningful progress. Without consistency, performance claims become unreliable, hindering innovation and eroding trust among researchers and practitioners. As such, maintaining consistent ML evaluation is fundamental for accurate, trustworthy, and impactful advancements across fields. Such research is worth undertaking as it addresses these critical challenges, promoting more reliable ML models, enhancing credibility, benefiting industries, and advancing society. Our selection of datasets and algorithms focused primarily on healthcare applications, especially medical imaging, which may restrict the generalizability of these findings to other domains. Nevertheless, these findings are sufficient to raise alarms and implications for a wide range of fields, as inconsistencies in evaluation metrics can impact any domain leveraging ML. Standardizing metrics would therefore benefit applications far beyond healthcare.

At present, there is no standardized approach to metric evaluation, posing challenges for achieving consistency across ML applications. To address these issues, our study discovered two main types of metric evaluation differences: Reporting Difference (RD) and Implementational Difference (ID). RD might occur when evaluation metrics are presented in different ways to meet various user needs. For example, someone in the ML community might configure a metric (e.g. Precision) to focus only on class zero, class one, or the average of both for specific reasons. As a result, users might only pay attention to the name of the metric component and not pay attention to how specifically it is set up. This can lead to confusion about how the metric is interpreted and presented. ID can arise when metric implementations vary across components due to programming mistakes, leading to discrepancies in computation results [5]. These differences further exacerbate the inconsistencies in ML metric evaluations, complicating model comparisons and undermining reliability.

In binary classification tasks, different evaluation metrics such as Accuracy, Balanced Accuracy, Cohen's Kappa, F-beta Score, MCC, G-mean, and Log Loss provided from different components were consistent across different programming languages. Furthermore, the evaluation metrics Precision, Recall, F1 Score, and Jaccard Index exhibit RD, indicating areas that may require further attention. Moreover, most of components were consistent except Caret, Yardstick, PyCM and TensorFlow. In fact, Caret and Yardstick calculate Precision, Recall, and F1 Score just for class 0, while PyCM provided for both classes 0 and 1. Other components, however, resulted only for class 1. For Jaccard Index, TensorFlow calculated the average values of two classes, while PyCM provided for both classes and others resulted for class1.

In multi-class classification tasks, different evaluation metrics such as Accuracy, Cohen's Kappa, and F-beta Score provided from different components look consistent across different programming languages. Furthermore, in evaluation metrics, Precision, Recall, F1 Score, Balanced Accuracy, and Jaccard Index indicated RD, while MCC demonstrates ID. Majority of components were consistent except TensorFlow, TorchMetrics, Scikit-Learn, Caret, Yardstick, PyCM, and Base R. TensorFlow provides two methods for calculating accuracy: tensorflow.keras.Accuracy and tensorflow.keras.CategoricalAccuracy, resulting in varying outcomes. Additionally, MCC can be calculated using two different formulas: one for binary types and another for multi-class scenarios. This can cause confusion and lead to errors.

TensorFlow and TorchMetrics provided micro values for Precision and Recall, while the other components returned values for all classes separately. Additionally, TorchMetrics provided the micro value for F1 Score, and the other components calculated it for all classes. For Balanced Accuracy, Matlab and Scikit-Learn provided the weighted value, while Yardstick provided the macro value. Results from Base R and Caret are different. For Jaccard Index most of components provided results for all classes, while TorchMetrics returned macro, and TensorFlow gives a different value. In MCC calculation, PyCM resulted in value for all classes and Base R gave macro, however, other libraries provided a completely different result.

In regression tasks, different evaluation metrics such as MAE, MSE, RMSE, MAPE, Explained Variance, Median AE, MSLE, and Huber provided from different components look consistent across different programming languages. Furthermore, in evaluation metrics $R^2$ and Tweedie Deviance exhibit RD and ID, respectively. In fact, for $R^2$, there are two approaches to calculate, Traditional $R^2$ only captures the linear relationship between variables (which is defined in Supplemental 4.2). The components such as Scikit-Learn, TorchMetrics, Base R, and MLmetrics use this formula to calculate $R^2$. Others employ a latter formula named Adjusted $R^2$ (it is defined in Supplemental 1.2.1) that can capture more complex relationships beyond linearity. When the model is linear, results are typically consistent. In Matlab, despite implementing both approaches, different results are often obtained. For Tweedie Deviance, the result from MetricsWeighted differs somewhat from those of Matlab and Scikit-Learn.

In clustering tasks, different evaluation metrics such as Davies-Bouldin Index, and Calinski-Harabasz Index provided from different components look consistent across different programming languages. Furthermore, in evaluation metrics Silhouette Score and WCSS exhibit ID. Scipy yielded a different result for the Silhouette Score in comparison of the other components. For WCSS, none of the components produced identical results. Since WCSS is directly related to the distances between data points and cluster centers, any changes in the cluster centers will impact the WCSS.

In correlation tasks, different evaluation metrics such as Pearson, Spearman, Kendall's Tau, Mutual Information, Distance Correlation, Percbend, Shepherd, and Partial Corr provided from different components look consistent across different programming languages. Furthermore, in evaluation metric Bicor exhibit ID. In fact, Pingouin and Matlab yielded different values for that.

In statistical test tasks, different evaluation metrics such as Paired t-test, Chi-Square Test, ANOVA, Kruskal-Wallis Test, Shapiro-Wilk Test, Welch's t-test, and Bartlett's test provided from different components look consistent across different programming languages. Furthermore, in evaluation metrics independent t-test, Kolmogorov-Smirnov Test, Mann-Whitney U Test, F-test and Levene's test exhibit ID. In fact, SymPy yielded a different result for p-value of independent t-test compared to other components. Matlab provided different results for p-value of Kolmogorov-Smirnov Test, statistics of Mann-Whitney U Test, p-value of F-test, and statistic & p-value of Levene's test. For p-value of Permutation test, none of the components produced identical results.

In 2D-segmentation tasks, different evaluation metrics such as Accuracy, Precision, and Recall provided from different components look consistent across different programming languages. Furthermore, in evaluation metric IoU, Hausdorff Distance and Dice Coefficient exhibit ID. In fact, TensorFlow and TorchMetrics have different results for metrics IoU and Dice Coefficinet, respectively, in comparison of the other components. Also, for Hausdorff Distance, TorchMetrics and MedPy resulted in the same value and differ from the others. In 3D-segmentation tasks, Accuracy evaluation metric provided from different components looks consistent across different programming languages. Furthermore, in evaluation metrics Precision, Recall, F1 Score, IoU, BF Score, Hausdorff Distance and Dice Coefficient seem to have ID.

In 2D-I2I translation tasks, different evaluation metrics such as MSE, and $R^2$ provided from different components look consistent across different programming languages. Furthermore, in evaluation metric PSNR and SSIM exhibit ID. In 3D-I2I translation tasks, different evaluation metrics such as MAE, MSE, and RMSE provided from different components look consistent across different programming languages. Furthermore, in evaluation metrics PSNR and SSIM seems to have ID, and $R^2$ has RD. In fact, Scikit-Image, TensorFlow and TorchMetrics resulted in different values. SSIM calculation for Scikit-Image is different from TensorFlow and TorchMetrics. Additionally, as explained in the regression section, we obtained different results due to the use of two different approaches in calculating $R^2$.

This study was prepared based on our knowledge and awareness of well-known metrics, and there may be other inconsistencies in metrics across different libraries, packages, and functions in programming languages Python, R, and Matlab. Attention should be given to works published using these metrics, and moving forward, greater caution should be exercised when using different metrics in various tasks. Additionally, based on the results obtained in this report, the need for a reliable standardization of various metrics across different programming languages seems essential, which could be considered as future work in this area.

This study has several limitations. First, while multiple datasets were used, their diversity and number may not fully capture all possible scenarios and variations. Second, some relevant metrics or components might have been unintentionally omitted, potentially affecting the analysis's completeness and accuracy. Additionally, the functionality of individual components and the accuracy of calculated metrics were not examined, limiting our ability to identify optimal performances and reliable metric computations. Future research will focus on verifying each component's correctness and the reliability of the metrics used. We also plan to develop a comprehensive library of evaluation metrics, ensuring their validity through rigorous validation processes. These steps will enhance the robustness and validity of our evaluation framework, leading to more reliable computational analyses.

## 5. CONCLUSION

We demonstrated a critical need for standardized approaches to the computation and reporting of ML evaluation metrics across different programming environments such as Python, R, and Matlab. The inconsistencies in both reporting and implementation of these metrics, identified as RD and ID, present challenges in ensuring the reliability and comparability of ML models across platforms. Furthermore, our findings indicate that relying solely on pre-defined metrics within these libraries is insufficient. Beyond the differences in reporting, we uncovered significant variations in how different metrics are implemented across libraries, packages, and functions in Python, R, and Matlab. These discrepancies highlight the importance of carefully validating the underlying methods used by these tools before employing them in critical ML model evaluations. A consistent approach is needed to mitigate these variations and ensure that the results generated by these tools are both reliable and comparable across different computational environments.

**DATA AND CODE AVAILABILITY**. All codes and Radiomics feature names are publicly shared at: https://github.com/MohammadRSalmanpour/Evaluation-Metrics-Evaluation

**ACKNOWLEDGMENT**. This study was supported by the Technological Virtual Collaboration Corporation Company (TECVICO CORP.) located in Vancouver, BC, Canada. We are also grateful to Eiliya Zanganeh, Sonya Falahati, and Amin Mousavi for their contributions in preparing some datasets such as clustering, segmentation tasks and facilitating I2I translation processes.

**CONFLICT OF INTEREST.** The authors declare no relevant conflict of interest.

## REFERENCE

[1] D. Sharma and N. Kumar, "A Review on Machine Learning Algorithms, Tasks and Applications," *International Journal of Advanced Research in Computer Engineering & Technology (IJARCET),* vol. 6, no. 10, 2017.

[2] G. Varoquaux and O. Colliot, "Evaluating machine learning models and their diagnostic value," in *Machine learning for brain disorders*, 2023, pp. 601-630.

[3] Y. Singh, P. K. Bhatia and O. Sangwan, "A Review of Studies on Machine Learning Techniques," *International Journal of Computer Science and Security,* vol. 1, no. 1, 2007.

[4] J. Feller, P. Finnegan and et al, "Developing open source software: a community-based analysis of research," in *Social Inclusion: Societal and Organizational Implications for Information Systems: IFIP TC8 WG8. 2 International Working Conference*, Springer US, 2006, pp. 261-278.

[5] B. Acko, B. Sluban and et al, "Performance metrics for testing statistical calculations in interlaboratory comparisons," *Advances in production engineering & management.,* vol. 9, no. 1, pp. 44-52, 2014.

[6] H. K. Bhuyan, V. Ravi and et al, "Disease analysis using machine learning approaches in healthcare system," *Health and Technology,* vol. 12, no. 5, pp. 987-1005, 2022.

[7] M. F. Dixon, I. Halperin and P. Bilokon, Machine Learning in Finance, New York, NY, USA: Springer International Publishing, 2020.

[8] E. Amigo´, J. Gonzalo and et al, "A comparison of extrinsic clustering evaluation metrics based on formal constraints," *Information retrieval,* vol. 12, pp. 461-486, 2009.

[9] P. Grabusts, "The choice of metrics for clustering algorithms," *ENVIRONMENT. TECHNOLOGIES. RESOURCES. Proceedings of the International Scientific and Practical Conference,* vol. 2, pp. 70-76, 2011.

[10] S. Emmons, S. Kobourov and et al, "Analysis of Network Clustering Algorithms," *PloS one,* vol. 11, no. 7, 2016.

[11] T. Abdel Aziz and A. Hanbury, "Metrics for evaluating 3D medical image segmentation: analysis, selection, and tool," *BMC medical imaging,* vol. 15, pp. 1-28, 2015.

[12] M. Dohmen, M. Klemens, I. Baltruschat, T. Truong and M. Lenga, "Similarity Metrics for MR Image-to-Image Translation," *arXiv preprint arXiv:2405.08431,* 2024.

[13] P. Yingxue, J. Lin and et al, "Image-to-image translation: Methods and applications," *IEEE Transactions on Multimedia,* vol. 24, pp. 3859-3881, 2021.

[14] D. Wu, L. Chen, Y. Zhou and B. Xu, "A metrics-based comparative study on object-oriented programming languages," *State Key Laboratory for Novel Software Technology at Nanjing University, Nanjing, China, DOI reference number 10,* 2015.

[15] M. Hossin and M. N. Sulaiman, "A review on evaluation metrics for data classification evaluations," *International journal of data mining & knowledge management process,* vol. 5, no. 2, p. 1, 2015.

[16] M. Salmanpour, M. Hosseinzadeh and et al, "Deep versus handcrafted tensor radiomics features: Application to survival prediction in head and neck cancer," in *EUROPEAN JOURNAL OF NUCLEAR MEDICINE AND MOLECULAR IMAGING*, 2022.

[17] M. Salmanpour, M. Hosseinzadeh and et al, "Robustness and Reproducibility of Radiomics Features from Fusions of PET-CT Images," in *Journal of Nuclear Medicine*, 2022.

[18] M. Salmanpour, I. Shiri and et al, "ViSERA: Visualized & Standardized Environment for Radiomics Analysis-A Shareable, Executable, and Reproducible Workflow Generator," in *2023 IEEE Nuclear Science Symposium, Medical Imaging Conference and International Symposium on Room-Temperature Semiconductor Detectors (NSS MIC RTSD)*, Vancouver, 2023.

[19] M. Salmanpour, M. Hosseinzadeh and et al, "Tensor Deep versus Radiomics Features: Lung Cancer Outcome Prediction using Hybrid Machine Learning Systems," in *Journal of Nuclear Medicine*, 2023.

[20] A. Gorji, M. Hosseinzadeh and et al, "Region-of-Interest and Handcrafted vs. Deep Radiomics Feature Comparisons for Survival Outcome Prediction: Application to Lung PET/CT Imaging," in *2023 IEEE Nuclear Science Symposium, Medical Imaging Conference and International Symposium on Room-Temperature Semiconductor Detectors (NSS MIC RTSD)*, 2023.

[21] M. Salmanpour, A. Mousavi and et al, "Do High-Performance Image-to-Image Translation Networks Enable the Discovery of Radiomic Features? Application to MRI Synthesis from Ultrasound in Prostate Cancer," in *International Workshop on Advances in Simplifying Medical Ultrasound*, 2024.

[22] M. Salmanpour, M. Hosseinzadeh and et al, "Prediction of TNM stage in head and neck cancer using hybrid machine learning systems and radiomics features," in *Medical Imaging 2022: Computer-Aided Diagnosis*, 2022.

[23] M. Salmanpour, A. Saberi and et al, "Optimal Feature Selection and Machine Learning for Prediction of Outcome in Parkinson's Disease," in *Journal of Nuclear Medicine*, 2020.

[24] M. Salmanpour, G. Hajianfar and et al, "Deep learning and machine learning techniques for automated PET/CT segmentation and survival prediction in head and neck cancer," in *3D Head and Neck Tumor Segmentation in PET/CT Challenge*, 2022.